\pdfoutput=1
\documentclass[11pt,a4paper]{article}
\usepackage[hyperref]{emnlp2020}
\usepackage{times}
\usepackage{latexsym}

\usepackage{color,soul}
\usepackage{tabularx}
\usepackage{graphicx}
\usepackage{dblfloatfix}
\usepackage{enumitem}
\usepackage{ textcomp }
\usepackage{todonotes}
\usepackage{subcaption}
\usepackage{times}
\usepackage{latexsym}
\usepackage{linguex}
\usepackage{amsmath}
\usepackage{amssymb}
\usepackage{booktabs}
\usepackage{linguex}
\usepackage{url}
\usepackage{array}
\usepackage{url}
\usepackage{lipsum}
\usepackage{multirow}
\usepackage{xcolor}

\usepackage[british]{babel}
\usepackage{hhline}
\usepackage{multirow}

\newcommand{\Hl}[2][\empty]{%
\ifx#1\empty
\else
\sethlcolor{#1}%
\fi
\hl{#2}}
\usepackage{soul,color}
\soulregister\Hl{7}
\soulregister\ref7

\usepackage{amssymb}
\usepackage{pifont}

\newcommand{\cmark}{\ding{51}}%
\newcommand{\xmark}{\ding{55}}%
\newcommand{\fem}[1]{\colorbox{pink}{#1}}
\newcommand{\masc}[1]{\colorbox{yellow}{#1}}

\usepackage{microtype}

\aclfinalcopy 
\def\aclpaperid{2739} 

\usepackage{fancyhdr}
\usepackage{geometry}
\newcommand\yg[1]{\textcolor{purple}{[YG: #1]}}
\newcommand\aj[1]{\textcolor{red}{[AJ: #1]}}
\newcommand\sr[1]{\textcolor{green!10!orange!90!}{[SR: #1]}}
\newcommand\rmv[1]{\textcolor{red!50!green!50!}{[RMV: #1]}}

\renewcommand\yg[1]{\textcolor{purple}{}}
\renewcommand\aj[1]{\textcolor{red}{}}
\renewcommand\sr[1]{\textcolor{green!10!orange!90!}{}}
\renewcommand\rmv[1]{\textcolor{red!50!green!50!}{}}

\title{Exposing Shallow Heuristics of Relation Extraction Models \\ with Challenge Data}

\author{Shachar Rosenman\textsuperscript{1} \;\;\; Alon Jacovi\textsuperscript{1} \;\;\; Yoav Goldberg\textsuperscript{1,2}\\
\textsuperscript{1}Computer Science Department, Bar Ilan University  \\
\textsuperscript{2}Allen Institute for Artificial Intelligence \\
  {\tt  \{shacharosn, alonjacovi, yoav.goldberg\}@gmail.com}
  }
\date{}

\begin{document}
\maketitle
\begin{abstract}
The process of collecting and annotating training data may introduce distribution artifacts which may limit the ability of models to learn correct generalization behavior. We identify failure modes of SOTA relation extraction (RE) models trained on TACRED, which we attribute to limitations in the data annotation process. We collect and annotate a challenge-set we call Challenging RE (CRE), based on naturally occurring corpus examples, to benchmark this behavior. Our experiments with four state-of-the-art RE models show that they have indeed adopted shallow heuristics that do not generalize to the challenge-set data. Further, we find that alternative question answering modeling performs significantly better than the SOTA models on the challenge-set, despite worse overall TACRED performance. By adding some of the challenge data as training examples, the performance of the model improves. Finally, we provide concrete suggestion on how to improve RE data collection to alleviate this behavior.

\end{abstract}

\thispagestyle{fancy}

\newcommand{\tableExamples}{
\begin{figure*}
\resizebox{\textwidth}{!}{
\begin{tabular}{lllc} \toprule

   Heuristic & Examples & Prediction  & Correct?\\ \midrule
  \multirow{3}{*}{Event} & (1) \masc{Edward} was born in York in \fem{1561}, the son of John, and his wife Mary.  &\emph{Birth date}  & \cmark  \\
    & (2) Edward was born in York in \fem{1561}, the son of \masc{John}, and his wife Mary.  &\emph{Birth date}   & \xmark \\
   & (3) Edward was born in York in \fem{1561}, the son of John, and his wife \masc{Mary}. &\emph{Birth date}  & \xmark \\
  \midrule
  \multirow{6}{*}{Event}  & (4) \masc{Loomis} is married to \fem{Hilary Mills}, who wrote a biography about Norman Mailer. &\emph{Spouse}  & \cmark  \\
    & (5) \masc{Loomis} is married to Hilary Mills, who wrote a biography about \fem{Norman Mailer}. &\emph{Spouse}  & \xmark \\
  & (6) \fem{Loomis} is married to \masc{Hilary Mills} , who wrote a biography about Norman Mailer. &\emph{Spouse}  & \cmark  \\
  & (7) \fem{Loomis} is married to Hilary Mills , who wrote a biography about \masc{Norman Mailer}. &\emph{Spouse}  & \xmark \\
  & (8) Loomis is married to \masc{Hilary Mills} , who wrote a biography about \fem{Norman Mailer}. &\emph{Spouse}  & \xmark \\
  & (9) Loomis is married to \fem{Hilary Mills} , who wrote a biography about \masc{Norman Mailer}. &\emph{Spouse}  & \xmark \\
  \midrule
  \multirow{3}{1.4cm}{Type \& Event} & (10) \masc{UCF} also has \fem{400} beds at the Rosen College Apartments Community, located on  Rosen College of Hospitality Management campus. &\emph{\# Members}   & \xmark \\
   & (11) UCF also has \fem{400} beds at the \masc{Rosen College Apartments Community}, located on the Rosen College of Hospitality Management campus. &\emph{\# Members}  & \xmark  \\
  & (12) UCF also has \fem{400} beds at the Rosen College Apartments Community, located on the \masc{Rosen College of Hospitality Management} campus. &\emph{\# Members}  & \xmark \\
  \bottomrule
  
\end{tabular}
}
\caption{CRE dataset instances illustrating the various heuristics and error types. ``\# Members'' refers to the number of human members or employees of an organization.} \label{fig:heur_examples}
\end{figure*}
}

\tableExamples

\section{Introduction}
\label{sec:intro}

In the relation extraction (RE) task, our goal is, given a set of sentences $s\in S$ to extract tuples $(s, e_1, e_2, r)$ where a relation $r\in R$ holds between $e_1$ and $e_2$ (entities that appear in the sentence $s$, each represented as a span over $s$). RE is often represented as relation classification (RC): given a triplet of $(s, e_1, e_2)$, determine which relation $r\in R$ holds between $e_1$ and $e_2$ in $s$, or indicate no-relation ($\emptyset$). This can be presented as a set of $|R|$ binary decision problems, $(s, e_1, e_2, r) \mapsto \{0,1\}$: return 1 for tuples for which the relation holds, and 0 otherwise. The reduction between RE and RC is clear: given a sentence, extract all entity pair candidates (given a NER system), and run the RC problem on each of them. 

Indeed, contemporary methods are all RC methods, and the popular TACRED large-scale relation extraction dataset is annotated for RC: each instance in the dataset is a triplet of $(s, e_1, e_2)$ and is associated with a label $r \in R\cup\{\emptyset\}$. Importantly, the annotation is \emph{non exhaustive}: not all $e_1$, $e_2$ pairs in the dataset are annotated (only 17.2\% of the entity pairs whose type match a TACRED relation are). While this saves a lot of annotation effort, as we show this also leads to sub-optimal behavior of the trained models, and hinders our ability to properly assess their real-world utility.

We show that state of the art models trained on TACRED are often ``right for the wrong reasons'' \cite{mccoy-etal-2019-right}: instead of learning to perform the intended task, they rely on shallow heuristics which are effective for solving many dataset instances, but which may fail on more challenging examples. In particular, we show two concrete heuristics: classifying based on \emph{entity types}, and classifying based on the \emph{existence of an event without linking the event to its arguments}. We show that while they are not well attested in the dev and test sets, these challenging examples do occur in practice. We introduce CRE (Challenging RE), a challenge set for quantifying and demonstrating the problem, and show that four SOTA RC models significantly fail on the challenge set. We release the challenge set to encourage future research on better models. \footnote{GitHub repository with data and code: \url{https://github.com/shacharosn/CRE}}

While we demonstrate the problem on TACRED, we stress that the model behaviors we expose are directly linked to the dataset construction procedure, and will likely occur in any dataset that is created in a similar fashion. We propose guidelines to help guide better datasets, and in particular better evaluation sets, in the future.

We also show that different modeling techniques may alleviate this problem: models trained for QA are better at linking events to their arguments. While performing worse on TACRED overall, they perform significantly better on the challenge set.

\section{Relation Classification Heuristics} \label{sec:problem}
\citet{mccoy-etal-2019-right} discusses the concept of ``model heuristics''---decision rules that are used by ML models to score high on a test set, but which are too simplistic to solve the underlying problem---and demonstrated such heuristics used by NLI models. In this work we demonstrate model heuristics used by TACRED-trained RC models.
Recall that a relation classification instance is $(s, e_1, e_2, r) \mapsto \{0,1\}$.

\noindent\textbf{\textit{Event} Heuristic:} Classify based on $(s,r)$. This heuristic ignores the entities altogether, acting as a classification model answering the question ``does the sentence attest the relation". This heuristic is of limited applicability, as many sentences attest more than a single related pair of entities.\\
\noindent\textbf{\textit{Type} Heuristic:} Classify based on $(type(e_1),$ $type(e_2), r)$, where $type(e)$ is the named-entity type of entity $e$. 
In a given dataset, a decision can be made based on the type of entities alone. For example of the 41 relations in the TACRED dataset, only the \emph{per:religion} relation is between a \emph{PERSON} and a \emph{RELIGION}. A model may learn to incorrectly rely on the types when making a decision, ignoring the sentence $s$ altogether.\footnote{Another, related, heuristic is ``classify based on $(e_1, e_2, r)$'', that is, based on prior knowledge about the entities. For a discussion of related problem see \cite{shwartz2020you}. This is beyond the scope of the current work.} 

Many type-pairs are compatible with multiple relations in a dataset, weakening the utility of this heuristic. However, for applicable type-pairs, it can be very effective.
For example, out of 21,284 Wikipedia sentences containing a PERSON name and a RELIGION name, 8,156 (38\%) were classified by a RoBERTA-based RC model as \textit{per:religion}. Manual inspection of a random sample of 100 of these, found that 42\% 
are 
false-positives.
\\
\noindent\textbf{\textit{Event+Type} Heuristic:} The \emph{event} and \emph{type} heuristics can be combined, by requiring the two decision rules $(s,r) \mapsto \{0,1\}$ and $(type(e_1), type(e_2), r) \mapsto \{0,1\}$ to hold. The resulting heuristic verifies that the sentence mentions the relation, and that the entity pairs of interest are type-compatible with the relation; it \emph{does not verify} that the entities are arguments of the relation.

We demonstrate that the event+type heuristic is a particularly strong one for relation-classification datasets, and is widely used by trained state-of-the-art relation classifiers. 

\section{Challenge Set}
 \label{sec:challenge-set}
Consider the \textit{date\_of\_birth} relation, that holds between a person${_{e_1}}$ and a year${_{e_2}}$, and the classification instance:\\[0.5em] 
\emph{[$_{e_1}$Steve Jobs] was born in California in [$_{e_2}$1955].}\\[0.5em]
A model making use of the event+type heuristic will correctly classify this relation.\footnote{Models that use the event heuristic on its own may predict \textit{birth\_location} instead, and models that rely on the type heuristic may predict \textit{date\_of\_death}.}
The heuristic is challenged by sentences that include multiple entities of an applicable type. For example:\\[0.5em]
\emph{[${_{e_1}}$Ed] was born in [$_{e_2}$1561], the son of John, a carpenter, and his wife Mary.}\\[0.5em]
A model relying solely on the event+type heuristic will correctly classify the above, but also incorrectly classify the following instances:\\[0.5em]
\emph{Ed was born in [$_{e_2}$1561], the son of [$_{e_1}$John], a carpenter, and his wife Mary.}\\[0.5em] 
\noindent\emph{Ed was born in York in [$_{e_2}$1561], the son of John, a carpenter, and his wife [$_{e_2}$Mary].}\\[0.5em]
While such sentences are frequent in practice, these cases are not represented in the dataset: in only 17.2\% of the sentences in TACRED more than a single pair is annotated (and only 3.46\% of the sentences have more than one different annotated labels in the sentence) \sr{This number is for sentences that have at least two different labels in the sentence}. Additionally, due to the data collection method \cite{zhang-etal-2017-position}, if a sentence includes a positive pair for a relation $r\in R$, it is significantly more likely that this pair will be chosen for annotation rather than a matching-type pair with a no-relation label. In other words, the data collection process leads to \emph{no-relation} labels between a pair of entities being assigned with very high probability to sentences in which other pairs in the sentence also do not adhere to a relation of interest.

As a result, for models trained on TACRED, the model is incentivized to learn to identify the existence of the relation in the sentence, irrespective of the arguments.  There is no signal in the data to incentivize the model to distinguish cases where the relation holds between the given arguments, from cases where the relation holds but between a different pair of arguments.
We expect the same to hold for \emph{any} large-scale RC dataset created in a similar manner.\footnote{Annotation of entity pairs, if the pairs are chosen at random or all pairs are annotated, carries a significant inefficiency in the number of annotations of related pairs vs. number of annotations of unrelated pairs. For this reason, RE annotation setups such as TACRED \cite{zhang-etal-2017-position} implement bias towards labeling pairs that are likely to be related, and thus, a no-relation label implies a higher likelihood of no relation event occurring in the sentence.}

\paragraph{Challenge Set Construction.}
We seek a benchmark to highlight RC models' susceptibility to the \textit{event+type} heuristic.
The benchmark takes the form of a challenge/contrast set; a collection of related examples that specialize in a specific failure case, meant only for evaluation and not training
\cite{kaushik2020-counterfactual,gardner2020-contrast-sets}.
In contrast to the NLI challenge set of \citet{mccoy-etal-2019-right}, in which challenge instances are based on synthetic examples, our benchmark is based on naturally occurring examples from real-world corpora. 

\paragraph{Methodology.} 
Coming up with a set of real-world sentences that demonstrates a failure mode is not easy. The main challenge is in identifying potential candidate sentences to pass to manual annotation. To identify such cases, we require an effective method for sampling a population that is likely to exhibit the behavior we are interested in.

We propose the following challenge-set creation methodology:
(1) use a strong \textit{seed-model} to perform large-scale noisy annotation; (2) identify suspicious cases in the model's output; (3) manually verify (annotate) suspicious cases. 

In stage (2), we identify suspicious cases by looking for sentences in which: (a) there are at least two entity-pairs of a NE type which is compatible to a TACRED relation (in most of the cases, the entity-pairs share one of the items, i.e. $(e_1,e_2), (e_1,e_3)$); and (b) these two pairs were assigned by the seed model to the same relation. Note that cases that satisfy condition (b) have, with high probability, at least one incorrect model prediction. On the other hand, given a strong seed model, there is also a high probability that one of the predictions is correct.

For our \textit{seed-model} we use a SOTA relation classification model which is based on fine-tuned SpanBERT \cite{joshi2019spanbert}, which we run over a large corpus of English Wikipedia sentences.\footnote{Additional details in the supplementary material.}

Out of the sentences that passed stage (2), we randomly sampled 100 sentences for each of the 30 relation predicted by the model. 
All instance entity pairs were manually labeled by two of the authors of this work, while closely adhering to the TACRED annotation guideline. For each instance, the annotators provide a \emph{binary} decision: does the entity pair in question adhere to relation $r$ (as predicted by the model) or not.\footnote{We chose to perform binary annotation, as we find it makes the annotation process faster and more accurate. As demonstrated by \citet{tacred-revisited}, multi-class relation labeling by crowd-workers lead to frequent annotation errors. We observed the same phenomena also with non-crowd workers.}

\noindent\textbf{The resulting challenge set (CRE)} has 3,000 distinct sentences, and 10,844 classification instances.
The dataset is arranged into 30 groups of 100 sentences, where each sentence is binary labeled for a given relation.
Example sentences from the challenge-set are given in Figure~\ref{fig:heur_examples}.
In 57\% of the sentences, there are at least two classification instances with conflicting labels, indicating the use of the event+type heuristic. On average there are 3.7 candidate entity-pairs per sentence. In 89.2\% of the sentences in the set, the entity-pairs share an argument.
Further details are available in the supplementary material.

\section{Evaluating RE Models}
By construction, the CRE dataset includes many instances that fail the seed RC model, which is based on fine-tuned SpanBERT \cite{joshi2019spanbert}. To verify that the behavior is consistent across models, we evaluate also RC models fine-tuned over other state-of-the-art LMs: BERT \cite{Devlin2019BERTPO}, RoBERTA \cite{Liu2019RoBERTaAR} and KnowBERT \cite{peters2019knowledge}. 
When evaluated on TACRED test set (Table~\ref{tab:TACRED_scores}), these models achieve SOTA scores. 

\begin{table}[t]
\centering

\scalebox{0.8}{
\begin{tabular}{l c c c}
\toprule
Model  & P & R & F$_1$ \\
\midrule
RC-SpanBERT & 70.8 & 70.9 & 70.8 \\
RC-BERT & 67.8 & 67.2 & 67.5 \\
RC-KnowBERT & 71.6 & 71.4 & 71.5 \\
RC-RoBERTa & 70.17 & 72.36 & 71.25 \\
\bottomrule
\end{tabular}}
\caption{Test results on TACRED.}
\label{tab:TACRED_scores}
\end{table}

We evaluate model's results on the CRE dataset in terms of accuracy.
We also report \textit{positive accuracy (\textbf{Acc$_+$})} (the accuracy on instances for which the relation hold; models that make use of the heuristic are expected to score high here) and likewise \textit{negative accuracy (\textbf{Acc$_-$})} (accuracy on instances in which the relation does not hold; models using the heuristic are expected to score low here). The models are consistently more accurate on the positive set then on the negative set showing that models struggle on cases where the heuristic make incorrect predictions.\footnote{CRE is binary labeled and relatively balanced between positive and negative examples, making accuracy a valid and natural metric. We chose to report \textbf{Acc$_+$} and \textbf{Acc$_-$} instead of the popular precision and recall because precision and recall emphasize the positive class, and do not tell the full story of the negative class (indeed, prec/rec do not involve the true-negative case), which is of interest to us. Using the \textbf{Acc$_{+/-}$} metric allows us to focus on both the positive and negative classes.}

A direct comparison to the state-of-the-art is difficult with these metrics alone.
To facilitate such a comparison, we also report precision, recall and F1 scores on \textbf{TACRED+Positive} (Table~\ref{tab:TACRED_and_challenge_POSITIVE}a), in which we add to the TACRED test set all the positive instances from the CRE dataset (easy cases), and on \textbf{TACRED+Negative} (Table~\ref{tab:TACRED_and_challenge_POSITIVE}b), in which we add the negative instances (hard cases).
All models benefit significantly from the positive setting, and are hurt significantly in the negative setting \textit{primarily in precision}, indicating that they do follow the heuristic on many instances: 

\textbf{the TACRED-trained models often classify based on the type of arguments and the existence of a relation in the text, without verifying that the entities are indeed the arguments of the relation.}

\begin{table}[t]
\centering
\scalebox{0.8}{
\begin{tabular}{l c c c}
\toprule
Model  & $Acc$ & $Acc_+$ & $Acc_-$ \\
\midrule
RC-SpanBERT & 63.5 & 89.7 & 42.5 \\
RC-BERT & 67.1 & 70.0 & 64.8 \\
RC-KnowBERT & 72.4 & 84.2 & 62.9 \\
RC-RoBERTa & 73.1 & 82.9 & 65.3 \\
\midrule
QA-SpanBERT & 75.5 & 71.5 & 78.7 \\
QA-BERT & 67.4 & 62.9 & 70.9 \\
QA-ALBERT & 75.3 & 71.5 & 78.8 \\
\bottomrule
\end{tabular}}
\caption{CRE accuracy for the RE and QA models. $Acc_+$ refers to accuracy on positive instances. $Acc_-$ refers to accuracy on negative instances. 
}
\label{tab:challenge-scores}
\end{table}

\begin{table}[t!]
\centering
\scalebox{0.85}{
\begin{tabular}{l c c c}
\toprule
Model  & P & R & F$_1$ \\
\midrule
\multicolumn{4}{c}{(a) TACRED + Positive}\\
\midrule
RC-SpanBERT & 88.2 & 79.3 & 83.5 \\
RC-BERT & 88.0 & 67.7 & 76.5 \\
RC-KnowBERT & 87.5 & 78.3 & 82.7 \\
RC-RoBERTa & 86.6 & 78.8 & 82.5 \\
\midrule
\multicolumn{4}{c}{(b) TACRED + Negative}\\
\midrule
RC-SpanBERT & 43.3 & 70.9 & 53.8 \\
RC-BERT & 42.0 & 64.0 & 50.7 \\
RC-KnowBERT & 43.9 & 71.6 & 54.4 \\
RC-RoBERTa & 43.6 & 72.7 & 54.5 \\
\bottomrule
\end{tabular}%
}
\caption{P/R/F$_1$ scores on TACRED test set + all positive instances from the CRE dataset (a), and + all negative instances (b).}
\label{tab:TACRED_and_challenge_POSITIVE}
\end{table}

\section{QA Models Perform Better}
The CRE dataset results indicate that RE-trained models systematically fail to link the provided relation arguments to the relation mention. We demonstrate that QA-trained models perform better in this respect. The QA models differ in both in their training data (SQuAD 2.0, \cite{rajpurkar-etal-2018-know}) and in their training objective (span-prediction, rather than classification). 
Inspired by \citet{levy-etal-2017-zero} we reduce RC instances into QA instances. We follow the reduction from \citet{cohen2020relation} between QA and binary relation classification which works by forming two questions for each relation instance, one for each argument. For example, for the relation instance pair \textit{(Mark, FB, founded)} we ask ``Who founded FB?'' and ``what did Mark found?''.\footnote{The question templates for each relation are defined manually. The full set, and additional details on the reduction method, are given in Appendix~\ref{sec:Detailed_QA}.} If the QA model answers either one of the questions with the correct span, we return 1 (relation holds), otherwise we return 0 (relation does not hold). 
We use three pre-trained SOTA models, fine-tuned on SQuAD 2.0: QA-SpanBERT, QA-BERT, and QA-AlBERT \cite{Lan2020ALBERTAL}.\\[0.5em]
\textbf{Results.} While the scores on the TACRED test set are unsurprisingly substantially worse (with F$_1$ of 59.1\%, 52.0\% and 61.4\%) than the TACRED trained models, they also perform better on the CRE dataset (Table~\ref{tab:challenge-scores}).

\textbf{QA-trained models pay more attention to the relation between an event and its arguments than RC-trained models.}

\section{Augmenting the Training Data with Challenge Set Examples}

We test the extent by which we can ``inoculate'' \cite{liu-etal-2019-inoculation} the relation extraction models by enhancing their training data with some examples from the challenge-set. We re-train each model on the TACRED dataset, which we augment with half of the challenge-set (5504 examples, 8\% of the size of the original TACRED training set). The other half of the challenge-set is used for evaluation.

\textbf{Results.} 
We begin by evaluating the inoculated models on the original TACRED evaluation set, establishing that this result in roughly the same scores, with small increases for most models (RC-SpanBERT: 71.0 F$_1$ (original: 70.8), RC-BERT: 69.9 F$_1$ (original:67.5), RC-KnowBERT: 72.1 F$_1$ (original:71.5), RC-RoBERTa: 70.8 F$_1$ (original:71.25)).

When evaluating on the CRE dataset examples, we see a large increase in performance for the inoculated dataset, as can be seen in Table~\ref{tab:acc_evaluate_part_of_challenge_set}. 
Compared to TACRED-only scores,
While we see a small and expected drop for \textbf{Acc$_{+}$}, it is accompanied by a very large improvement on \textbf{Acc$_{-}$}. However, while accuracies improve, there is still a wide gap from perfect accuracy.


\begin{table}[t!]
\centering
\scalebox{0.85}{
\begin{tabular}{l c c c}
\toprule
Model  & $Acc$ & $Acc_+$ & $Acc_-$ \\
\midrule
\multicolumn{4}{c}{(a) Trained on TACRED}\\
\midrule
RC-SpanBERT & 62.8 & 89.5 & 41.6 \\
RC-BERT & 65.8 & 68.4 & 63.7 \\
RC-KnowBERT & 71.6 & 83.0 & 62.5 \\
RC-RoBERTa & 75.5 & 85.4 & 68.0 \\
\midrule
\multicolumn{4}{c}{(b) Trained on TACRED + half CRE }\\
\midrule
RC-SpanBERT & 84.4 & 85.7 & 83.4 \\
RC-BERT & 78.7 & 86.1 & 72.7 \\
RC-KnowBERT & 82.4 & 81.9 & 82.7 \\
RC-RoBERTa & 83.0 & 83.4 & 82.6 \\
\bottomrule
\end{tabular}
}
\caption{Acc/$Acc_+$/$Acc_-$ scores on half of the CRE dataset, models trained on TACRED training set (a), models trained on TACRED training set with examples from the second half of the CRE dataset (b).}
\label{tab:acc_evaluate_part_of_challenge_set}
\end{table}

\section{Discussion and Conclusion}
We created a challenge dataset demonstrating the tendency of TACRED-trained models to classify using an event+type heuristic that fails to connect the relation and its arguments. QA-trained models are less susceptible to this behavior. 
Continuing \citet{gardner2020-contrast-sets}, we conclude that challenge sets are an effective tool of benchmarking against shallow heuristics, not only of models and systems, but also of data collection methodologies.

We suggest the following \textbf{recommendation for future RE data collection}: \emph{evaluation sets} should be exhaustive, and contain \emph{all} relevant entity pairs. Ideally, the same should apply also to \emph{training sets}. If impractical, the data should at least attempt to exhaustively annotate \emph{confusion-sets}: if a certain entity-pair is annotated in a sentence, all other pairs of the same entity-types in the sentence should also be annotated.

\section*{Acknowledgements}
We thank Amir David Nissan Cohen for fruitful discussion and for sharing code.
This project has received funding from the Europoean Research Council (ERC) under the Europoean Union's Horizon 2020 research and innovation programme, grant agreement No. 802774 (iEXTRACT).

\bibliography{emnlp2020}
\bibliographystyle{acl_natbib}

\newcommand{\tableResultsREonChallenge}{
\begin{table*}[b]
\resizebox{\textwidth}{!}{
\begin{tabular}{lcccccccccc} \toprule

   Model  & P & R & F$_1$ & TP & FP & TN & FN & Pos\_acc & Neg\_acc & Total\_acc\\ \midrule
  RC-SpanBERT & 55.6 & 89.7 & \multicolumn{1}{c|}{68.6} & 39.9 & 31.8 & 23.6 & \multicolumn{1}{c|}{4.5} & 89.7 & 42.5 & 63.5\\
  RC-BERT & 61.0 & 70.0 & \multicolumn{1}{c|}{65.5} & 31.1 & 19.5 & 36.0 & \multicolumn{1}{c|}{13.3} & 70.0 & 64.8 & 67.1 \\
  RC-KnowBERT & 64.5 & 84.1 & \multicolumn{1}{c|}{73.0} & 37.4 & 20.5 & 34.9 & \multicolumn{1}{c|}{7.0} & 84.2 & 62.9 & 72.4\\
  RC-RoBERTa & 65.7 & 82.8 & \multicolumn{1}{c|}{73.3} & 36.8 & 19.2 & 36.3 & \multicolumn{1}{c|}{7.6} & 82.9 & 65.3 & 73.1\\
  \midrule
  QA-SpanBERT & 72.9 & 71.5 & \multicolumn{1}{c|}{72.2} & 31.8 & 11.7 & 43.7 & \multicolumn{1}{c|}{12.6} & 71.5 & 78.7 & 75.5\\
  QA-BERT & 63.4 & 62.9 & \multicolumn{1}{c|}{63.0} & 28.0 & 16.1 & 39.4 & \multicolumn{1}{c|}{16.4} & 62.9 & 70.9 & 67.4\\
  QA-ALBERT & 72.6 & 71.5 & \multicolumn{1}{c|}{72.0} & 31.8 & 12.0 & 43.5 & \multicolumn{1}{c|}{12.6} & 71.5 & 78.3 & 75.3\\
  \bottomrule
  
\end{tabular}
}
\caption{Results of models on the CRE dataset.
} \label{tab:results_on_challenge_set}
\end{table*}
}
\tableResultsREonChallenge

\newpage

\appendix

\section{Challenge Set}

We use SpanBERT as a seed model, SpanBert is a recent state-of-the-art model uses pre-trained language model fine-tuned to the RC task which uses a bidirectional LM pre-trained on SpanBERT
We run the seed model over a large corpus of English Wikipedia sentences in which contains more than 10 million sentences.
Table ~\ref{tab:challenge_set_stats} contain all the relations in the challenge set with number of positive and negative instances per relation.
The challenge set is relatively balanced in terms of positive examples (44\%) and negative examples (56\%), and the classification is binary, so it is possible to use accuracy as evaluation method. The average length of a sentence is 28.7 tokens.
Crucially, there are no sentences with less than two pair of entities.

\section{Experiment Details}

\paragraph{RC models}
We evaluate four RC models fine-tuned on TACRED:
\textbf{SpanBERT} \cite{joshi2019spanbert} extends BERT model by pre-trained on a span-level;
\textbf{KnowBERT} \cite{peters2019knowledge} which integrates knowledge bases into BERT,  we use KnowBERT-W+W, where the knowledge comes from joint entity linking and language modelling on Wikipedia and WordNet;
\textbf{RoBERTa} \cite{Liu2019RoBERTaAR}, we use the same baseline from \cite{wang2020k};
\textbf{BERT} \cite{Devlin2019BERTPO} we use the same model from \cite{joshi2019spanbert} for BERT base.
Table~\ref{tab:TACRED_scores} contain the performance of the four models on TACRED.


\paragraph{QA models}
We use three pre-trained SOTA models, fine-tuned on SQuAD 2.0:
\textbf{QA-SpanBERT}: SpanBERT \cite{joshi2019spanbert}  we use a pre-trained model \cite{joshi2019spanbert} fine-tuned on SQuAD 2.0 for QA ($F_{1}$: 88.7). 
\textbf{QA-BERT}: BERT \cite{Devlin2019BERTPO}, we use a pre-trained model \cite{joshi2019spanbert} fine-tuned on SQuAD 2.0 for QA. ($F_{1}$: 83.3)
\textbf{QA-ALBERT}: ALBERT xxlarge \cite{Lan2020ALBERTAL}, we use a pre-trained model \cite{elgeish2020gestalt} fine-tuned on SQuAD 2.0 for QA. ($F_{1}$: 88.9)

\section{How we evaluate TACRED + pos/neg}

In order to evaluate TACRED + Positive and TACRED+Negative we presented them as binary decision problem as explained in section \ref{sec:intro}.
We simulate ACRED + pos/neg in which for each relation $r\in R$ we create a set that contain all the examples  $(s, e_1, e_2, r) \mapsto \{0,1\}$ , which the arguments types $e_{1}$ and $e_{2}$ match the relation $r$.
We evaluate each set of relation $r$ separately, and  we report the results as micro-averaged F1 scores.

\section{Detailed QA reduction}
\label{sec:Detailed_QA}
Table ~\ref{tab:question_examples} contain the templates for the
questions, each relation have two questions, a question for the head entity and a question for the tail entity, we use this templates to reduce the relation classification task.

\section{Detailed Results}
Table~\ref{tab:results_on_challenge_set}, \ref{detailed} shows the full results of all the models on the challenge set, and table~\ref{detailed} shows the full results of all the models on  \textbf{TACRED+Positive} and \textbf{TACRED+Negative}, $\Delta$ denotes the difference in performance between them, All models preformed a big difference in precision between the positive and negative examples.


\newcommand{\tableResultsREonTACRED}{
\begin{table*}[b]
\resizebox{\textwidth}{!}{
\begin{tabular}{lcccccccccc} \toprule

   Model  & P & R & F$_1$ & TP & FP & TN & FN & Pos\_acc & Neg\_acc & Total\_acc\\ \midrule
  RC-SpanBERT & 70.9 & 72.9 & \multicolumn{1}{c|}{71.9} & 4.3 & 1.5 & 92.3 & \multicolumn{1}{c|}{1.7} & 70.9 & 98.3 & 96.6\\
  RC-BERT & 64.0 & 73.6 & \multicolumn{1}{c|}{68.4} & 3.8 & 1.3 & 92.5 & \multicolumn{1}{c|}{2.1} & 64.0 & 98.5 & 96.4 \\
  RC-KnowBERT & 71.6 & 72.1 & \multicolumn{1}{c|}{71.8} & 4.3 & 1.6 & 92.2 & \multicolumn{1}{c|}{1.7} & 71.6 & 98.2 & 96.6\\
  RC-RoBERTa & 72.7 & 70.6 & \multicolumn{1}{c|}{71.6} & 4.4 & 1.8 & 9.2 & \multicolumn{1}{c|}{1.6} & 72.7 & 98.0 & 96.5\\
  
  \midrule
  QA-SpanBERT & 60.5 & 57.8 & \multicolumn{1}{c|}{59.1} & 3.6 & 2.6 & 91.2 & \multicolumn{1}{c|}{2.3} & 60.0 & 97.1 & 94.9\\
  QA-BERT & 54.5 & 50.1 & \multicolumn{1}{c|}{52.0} & 3.3 & 3.0 & 90.6 & \multicolumn{1}{c|}{2.7} & 54.5 & 96.4 & 93.0\\
  QA-ALBERT & 62.8 & 60.1 & \multicolumn{1}{c|}{61.4} & 3.8 & 2.5 & 91.4 & \multicolumn{1}{c|}{2.2} & 62.8 & 97.3 & 95.2\\
  \bottomrule
  
\end{tabular}
}
\caption{Results of models on simulated TACRED.
} \label{tab:heur_examplesss}
\end{table*}
}

\begin{table*}[b]
  \centering
  \renewcommand{\arraystretch}{1.2}
  \begin{tabular}{l c c c c c c c c c} \toprule
    \multirow{2}{4cm}{\textbf{Models}} & \multicolumn{3}{c}{\textbf{+ positive}} & \multicolumn{3}{c}{\textbf{+ negative}} & \multicolumn{3}{c}{\textbf{Difference}}\\
    & P & R & F$_1$ & P & R & F$_1$ & $\Delta_P$ & $\Delta_R$ & $\Delta_{F_1}$ \\
    \midrule
    \multicolumn{1}{l|}{RC-SpanBERT} & 88.2 & 79.3 &   \multicolumn{1}{l|}{83.5}  & 43.3 & 70.9 & \multicolumn{1}{l|}{53.8} & 44.9 & 8.4 & 29.7 \\ 
    \multicolumn{1}{l|}{RC-BERT} & 88.0 & 67.7  & \multicolumn{1}{l|}{76.5} & 42.0 & 64.0 & \multicolumn{1}{l|}{50.7} & 46.0 & 3.7 & 25.8\\ 
    \multicolumn{1}{l|}{RC-KnowBERT} & 87.5 & 78.3  & \multicolumn{1}{l|}{82.7} & 43.9 & 71.6 & \multicolumn{1}{l|}{54.4} & 43.6 & 6.7 & 28.3\\ 
    \multicolumn{1}{l|}{RC-RoBERTa} & 86.6 & 78.8 & \multicolumn{1}{l|}{82.5} & 43.6 & 72.7 &  \multicolumn{1}{l|}{54.5 }  & 43.0 & 6.1 & 28.0\\ \midrule
    \multicolumn{1}{l|}{QA-SpanBERT} & 70.6 & 67.1  & \multicolumn{1}{l|}{68.8} & 36.0 & 60.7 & \multicolumn{1}{l|}{45.2 }  & 34.6 & 6.4 & 23.6\\ 
    \multicolumn{1}{l|}{QA-BERT} & 66.3 & 63.7 & \multicolumn{1}{l|}{65.0} & 32.7 & 64.9 & \multicolumn{1}{l|}{43.4}  & 33.6 & -1.2 & 21.6\\ 
    \multicolumn{1}{l|}{QA-ALBERT} & 80.5 & 70.0 & \multicolumn{1}{l|}{74.9} & 41.9 & 63.0 & \multicolumn{1}{l|}{50.3}  & 38.6 & 7.0 & 24.6\\ 
    \bottomrule

  \end{tabular}
  \caption{RE and QA models on TACRED + negative, TACRED + positive, and the differences between them.} \label{detailed}
\end{table*}

\newcommand{\tableTemplate}{
\begin{table*}
\resizebox{\textwidth}{!}{
\begin{tabular}{lll}
\toprule
\textbf{Relation} & \textbf{Question 1} & \textbf{Question 2} \\ \midrule
\emph{org:founded\_by} & Who founded $e_{1}$? &  What did $e_{2}$ found? \\
\emph{per:employee\_of} & Where does $e_{1}$ work? &  Who is an employee of $e_{2}$? \\
\emph{per:title} & What is $e_{1}$'s title? &  Who has the title $e_{2}$? \\
\emph{per:age} & What is $e_{1}$'s age? &  Whose age is $e_{2}$? \\
\emph{per:date\_of\_birth} & When was $e_{1}$ born? &  What is $e_{2}$'s date of birth? \\
\emph{org:top\_members/employees} & Who are the top members of the organization $e_{1}$?&  $e_{2}$ is  a top member of which organization? \\
\emph{org:country\_of\_headquarters} & in what country the headquarters of $e_{1}$ is? & What is the country that $e_{2}$'s headquarters located in? \\
\emph{per:parents} & Who are the parents of $e_{1}$? & Who are the parents of $e_{2}$? \\
\emph{per:countries\_of\_residence} & What country does $e_{2}$ resides in? & In what country does $e_{2}$ live \\
\emph{per:children} & Who are the children of $e_{1}$? & Who are $e_{2}$'s children? \\
\emph{org:alternate\_names} & What is the alternative name of the organization $e_{2}$? & What are other names for $e_{2}$? \\
\emph{per:charges} & What are the charges of $e_{1}$? & What is $e_{2}$ charged with? \\
\emph{per:cities\_of\_residence} & What city does $e_{2}$ resides in? & What is $e_{2}$'s cities of residence? \\
\emph{per:origin} & What is $e_{1}$ origin? & Where's $e_{2}$'s origin? \\
\emph{per:siblings} & Who is the sibling of $e_{1}$? & Who is the brother of {}? \\
\emph{per:alternate\_names} & What is the alternative name of $e_{1}$? & {} is another name for which person? \\
\emph{org:website} & What is the URL of $e_{1}$ & what is the website address of {}? \\
\emph{per:religion} & What is the religion of $e_{1}$? & What religion does $e_{2}$ believe in? \\
\emph{per:stateorprovince\_of\_death} & Where did $e_{1}$ died? & What is the place where $e_{2}$ died? \\
\emph{org:parents} & What organization is the parent organization of $e_{1}$? & What organization is the parent organization of $e_{2}$? \\
\emph{org:subsidiaries} & What organization is the child organization of $e_{1}$? & What organization is the child subsidiaries of $e_{2}$? \\
\emph{per:other\_family} & Who are family of $e_{1}$? & ho are family of $e_{2}$? \\
\emph{per:stateorprovinces\_of\_residence} & What is the state of residence of $e_{1}$? & Where is $e_{2}$'s place of residence? \\
\emph{org:members} & Who is a member of the organization $e_{1}$? & What organization $e_{2}$ is member of? \\
\emph{per:cause\_of\_death} & How did $e_{1}$ died? & What is $e_{2}$'s cause of death? \\
\emph{org:member\_of} & What is the group the organization $e_{1}$ is member of? &What organization is a member of $e_{2}$?  \\
\emph{org:number\_of\_employees/members} & How many members does $e_{1}$ have? & What is the number of members of $e_{2}$? \\
\emph{per:country\_of\_birth} & In what country was $e_{1}$ born & hat is $e_{2}$'s country of birth? \\
\emph{org:shareholders} & Who hold shares of $e_{1}$? & Who are $e_{2}$'s shareholders? \\
\emph{org:stateorprovince\_of\_headquarters} & What is the state or province of the headquarters of $e_{1}$? & Where is the state or province of the headquarters of $e_{2}$? \\
\emph{per:city\_of\_death} & In what city did $e_{1}$ died? & What is $e_{2}$'s city of death? \\
\emph{per:city\_of\_birth} & In what city was $e_{1}$ born? & What is $e_{2}$'s city of birh? \\
\emph{per:spouse} & Who is the spouse of $e_{1}$? & Who is the spouse of $e_{2}$? \\
\emph{org:city\_of\_headquarters} & Where are the headquarters of $e_{1}$? & What is $e_{2}$'s city of headquarters? \\
\emph{per:date\_of\_death} & When did $e_{1}$ die? & What is $e_{2}$'s date of death? \\
\emph{per:schools\_attended} & Which schools did {} attend? & What school did $e_{2}$ attend? \\
\emph{org:political/religious\_affiliation} & What is $e_{1}$ political or religious affiliation? & What religion does $e_{1}$ organization belong to? \\
\emph{per:country\_of\_death} & Where did $e_{1}$ die? & What $e_{2}$'s country of death? \\
\emph{org:founded} & When was $e_{1}$ founded? & What date did $e_{2}$ establish? \\
\emph{per:stateorprovince\_of\_birth} & In what state was $e_{1}$ born? & What is $e_{2}$'s country of birth? \\
\emph{per:city\_of\_birth} & Where was $e_{1}$ born? & Who was born in $e_{2}$? \\
\emph{org:dissolved} & When was $e_{1}$ dissolved? & What date did $e_{2}$ dissolved? \\

\bottomrule
\end{tabular}%
}
\caption{Templates for the  questions, for each relation two questions are defined.}
\label{tab:question_examples}
\end{table*}
}

\tableTemplate

\newcommand{\tableChallengeSetStats}{
\begin{table*}
\centering
\begin{tabular}{l c c}
\toprule
Relation & Positive & Negative \\
\midrule
\emph{org:founded\_by} & 214 & 179 \\
\emph{per:age} & 114 & 186 \\
\emph{per:date\_of\_birth} & 150 & 158 \\
\emph{org:founded}  & 123 & 190\\
\emph{per:schools\_attended} & 193 & 175\\
\emph{per:employee\_of} & 209 & 120\\
\emph{org:country\_of\_headquarters} & 136 & 120\\
\emph{per:alternate\_names} & 390 & 210\\
\emph{per:children} & 141 & 459\\
\emph{per:city\_of\_birth} & 108 & 148\\
\emph{per:city\_of\_death} & 119 & 103\\
\emph{per:date\_of\_death} & 112 & 153\\
\emph{per:religion} & 174 & 77 \\
\emph{per:spouse} & 212 & 388\\
\emph{org:alternate\_names} & 357 & 243\\
\emph{per:title} & 126 & 111\\
\emph{org:parents} & 141 & 459\\
\emph{per:other\_family} & 332 & 268\\
\emph{per:stateorprovince\_of\_birth} & 121 & 105\\
\emph{org:political/religious\_affiliation} & 119 & 101\\
\emph{per:siblings} & 232 & 368\\
\emph{per:origin} & 153 & 129\\
\emph{per:cities\_of\_residence} & 102 & 157\\
\emph{org:city\_of\_headquarters} & 129 & 121\\
\emph{per:countries\_of\_residence} & 95 & 152\\
\emph{per:parents} & 142 & 458\\
\emph{per:stateorprovinces\_of\_residence} & 110 & 154\\
\emph{org:top\_members/employees} & 89 & 189\\
\emph{org:stateorprovince\_of\_headquarters} & 135 & 86 \\
\emph{org:number\_of\_employees/members} & 41 & 258\\
\midrule
Total & 4819 & 6025\\
\bottomrule
\end{tabular}
\caption{CRE dataset number of positive and negative instances per relation}
\label{tab:challenge_set_stats}
\end{table*}
}
\tableChallengeSetStats





\end{document}